\pdfoutput=1

\documentclass[11pt]{article}

\usepackage{acl}

\usepackage{times}
\usepackage{latexsym}

\usepackage[T1]{fontenc}

\usepackage[utf8]{inputenc}

\usepackage{microtype}

\usepackage{inconsolata}
\usepackage{graphicx}
\graphicspath{ {./images/} } 

\usepackage{amsmath} 
\usepackage{multirow}

\title{PEDAL: Enhancing Greedy Decoding with Large Language Models using Diverse Exemplars }

\author{Sumanth Prabhu \\
  \texttt{sumanth@parsimoai.com}}

\begin{document}
\maketitle
\begin{abstract}
Self-ensembling techniques with diverse reasoning paths such as Self-Consistency have demonstrated remarkable performance gains in text generation with Large Language Models (LLMs). However, such techniques depend on the availability of an accurate answer extraction process to aggregate across multiple outputs. Moreover, they acquire higher inference cost, in comparison to Greedy Decoding, due to generation of relatively higher number of output tokens. Research has shown that the free form text outputs from Self-Consistency can be aggregated reliably using LLMs to produce the final output.  Additionally, recent advancements in LLM inference have demonstrated that usage of diverse exemplars in prompts have the ability to induce diversity in the LLM outputs. Such proven techniques can be easily extended to self-ensembling based approaches to achieve enhanced results in text generation. In this paper, we introduce PEDAL (Prompts based on Exemplar Diversity Aggregated using LLMs), a hybrid self-ensembling approach, that combines the strengths of diverse exemplar based prompts and LLM based aggregation to achieve improvement in overall performance. On the publicly available SVAMP and ARC datasets, our experiments reveal that PEDAL can achieve better accuracy than Greedy Decoding based strategies with lower inference cost compared to Self Consistency based approaches.     

\end{abstract}

\section{Introduction}
Large Language Models (LLMs)~\cite{brown2020GPT3,Raffel2020T5,Chowdhery2022PaLMSL,Touvron2023LLaMAOA} have been proven to show remarkable performance in a wide range of Natural Language Understanding tasks~\cite{zhao2023surveylargelanguagemodels} as a result of their outstanding reasoning capabilities~\cite{Wei2022ChainOT,Zhou2022LeasttoMostPE}.
However, they still rely on carefully designed prompts to achieve optimal performance~\cite{khattab2023dspycompilingdeclarativelanguage,fernando2023promptbreederselfreferentialselfimprovementprompt}.  To realize further improvement in LLM reasoning, \cite{Wang2022SelfConsistencyIC}~proposed a self-ensembling technique termed ``Self-Consistency''(SC) where diverse ``Chain-of-Thought''(CoT)~\cite{Wei2022ChainOT} reasoning paths were generated and then aggregated to construct an accurate and reliable response. This approach has been successfully extended to various use-cases such as LLM hallucination detection~\cite{chen2024insidellmsinternalstates}, medicine\cite{zhou2024LLMSurveyMedocine} and code generation~\cite{huang2024enhancinglargelanguagemodels}.

While SC based approaches can significantly improve the robustness of LLM outputs, one of their common drawbacks is that they perform best on a fixed answer set~\cite{Wang2022SelfConsistencyIC} or rely on training custom aggregation methods to measure consistency across multiple text outputs. To address this, \cite{Chen2023UniversalSF}~proposed ``Universal Self Consistency''(USC), an extension of SC, that aggregated the text outputs by re-invoking the LLM. Essentially, USC prompted the LLM to select the most consistent response among the different candidate answers generated by SC and demonstrated that it can achieve improved performance. However, this still leaves us with another drawback of SC which is the cost involved in generating the outputs. Concretely, SC involves generating long and diverse reasoning paths which results in a higher number of output tokens compared to Greedy Decoding based approaches. The cost of output token generation with LLMs is typically more than input token processing due to the difference in the number of forward passes~\cite{shazeer2019fasttransformerdecodingwritehead,Chng2024OutputTokenGenCost} resulting in a higher inference cost with SC.

\cite{li-etal-2023-diverse-prompt}~experimented with usage of diverse exemplars in the LLM prompts and combined them with diverse reasoning paths in SC to achieve more accurate results in text generation. We observe that if we leverage diverse exemplars with Greedy Decoding for text generation and aggregate the responses as in USC, we achieve better performance than traditional Greedy Decoding in terms of accuracy while also achieving lower cost of inference in comparison to SC based approaches. 
 
In this paper, we present a hybrid self-ensembling approach, PEDAL(\textbf{P}rompts based on \textbf{E}xemplar \textbf{D}iversity \textbf{A}ggregated using an \textbf{L}LM), that offers a trade-off between the Greedy Decoding and SC in terms of accuracy and cost efficiency. We leverage diverse exemplars in LLM prompts to generate multiple candidate responses using Greedy Decoding and then aggregate them using an LLM to generate the final response. On two publicly available datasets, we demonstrate that PEDAL achieves better accuracy than Greedy Decoding based strategies and offers lower cost in inference compared to SC based strategies. 

Rest of the paper is organized as follows: In
Section~\ref{sec:related-work}, we describe previous work for solving similar problems. Section~\ref{sec:methodology} explains our proposed strategy in detail followed by Section~\ref{sec:experiments} where we describe the data and the experiment settings to validate PEDAL. We then present our results and analyses in Section~\ref{sec:analysis}. Finally, in Section~\ref{sec:conclusion}, we summarize our findings and discuss potential future work.

\section{Related Work}\label{sec:related-work}
LLMs have been widely studied and applied in a variety of tasks including code generation~\cite{zheng2024surveylargelanguagemodelsCode}, finance~\cite{li2024largelanguagemodelsfinance}, law~\cite{yu2022legalpromptingteachinglanguage} and so on. However, none of the LLMs seem to consistently outperform the rest of the models across all tasks~\cite{jiang2023llmblenderensemblinglargelanguage}. This led to exploring ensembling approaches with LLMs. Research focused on Prompt Chaining~\cite{Chase_LangChain_2022}, Fusion~\cite{li2023deepmodelfusionsurvey}, Mixture of Experts~\cite{cai2024surveymixtureexperts} and many more have shown promising results in combining LLMs to enhance the overall performance.

\subsection{Self Ensembling Strategies}

\cite{long2023largelanguagemodelguided,yao2023treethoughtsdeliberateproblem}~generalized CoT to organize language model generated
“thoughts” into a tree structure for solution
search. However, similar to~\cite{Wang2022SelfConsistencyIC}, they rely on custom aggregation methods to construct the final output. \cite{Chen2023UniversalSF}~addressed this issue by leveraging LLMs to perform majority consensus based aggregation without any specific model fine-tuning. In our work, we leverage a similar strategy to aggregate multiple candidates with a focus on the impact of using diverse LLM prompts as opposed to diverse reasoning paths.

\subsection{Prompt Ensembling Strategies}
With the advent of LLMs, lot of research focused on developing effective prompting techniques~\cite{bach-etal-2022,lu-etal-2022-fantastically-pe} that have been extended by multiple prompt ensembling techniques~\cite{zhang2023preferpromptensemblelearning,pitis2023boostedpromptensembleslarge} to achieve further improvement. \cite{singh-etal-2023-tree}~built a decision tree of prompts that links multiple LM calls to solve a task. \cite{arora2022askanythingsimplestrategy}~used multiple prompt templates to reformat few-shot example inputs into an open ended question-answering format and then leverage Weak Supervision~\cite{WeakSupervisionSnorkel2017} to aggregate the LLM predictions. \cite{Hou2023PromptBoosting}~applied AdaBoost~\cite{schapire2013explaining} algorithm over a pre-defined prompt set for text classification by pairing prompts with the corresponding output distribution to construct a large pool of weak learners. \cite{li-etal-2023-diverse-prompt}~enhanced SC with diverse prompts by randomly selecting different exemplars for prompt construction, followed by sampling reasoning paths for each such prompt and then scoring the quality of each reasoning path using a custom trained model. While our work also leverages a similar prompt construction strategy, we aggregate the predictions without relying on explicitly training a task-specific model. Additionally, we focus on leveraging such prompt based strategies to reduce LLM inference cost rather than enhancing SC based approaches.

\begin{figure*}[t]
    \centering
    \includegraphics[width=14cm, trim={0cm 0cm 0cm 0cm}, ]{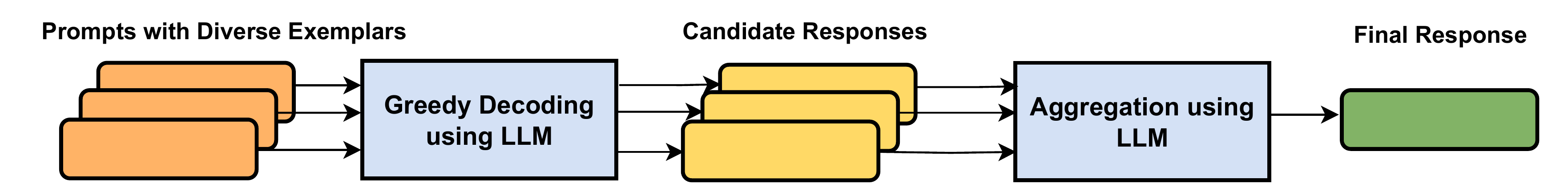}
    \caption{High level overview of PEDAL (\textbf{P}rompts based on \textbf{E}xemplar \textbf{D}iversity \textbf{A}ggregated using an \textbf{L}LM)}
    \label{fig:block-diagram}
\end{figure*}

\subsection{LLM Inference Cost}
To solve the problem of inference cost, researchers have commonly explored model compression techniques~\cite{zhu2024surveymodelcompressionlarge} such as model quantization~\cite{Jacob_2018_CVPR}, model pruning~\cite{cheng2024surveydeepneuralnetwork} and model distillation~\cite{Gou_2021} aimed at reducing the size of the model without hurting the performance significantly. \cite{shazeer2019fasttransformerdecodingwritehead}~proposed sharing keys and values across all of the different attention heads in the transformer architecture, thus, reducing the memory bandwidth requirements of incremental decoding. \cite{wu2024paralleldecodinghiddentransfer}~explored decoding multiple successive tokens simultaneously in a single forward pass to reduce the inference time. FrugalGPT~\cite{chen2023frugalgptuselargelanguage} proposed a cascade of LMs that stops when an intermediate output is considered reliable, resulting in better computational efficiency. In our work, we focus on reducing the number of output tokens during LLM inference in comparison to SC while achieving better accuracy than Greedy Decoding.

\section{Methodology}\label{sec:methodology}

Figure~\ref{fig:block-diagram} shows the high level overview of our proposed system. The LLM generates multiple candidate responses using Greedy Decoding with prompts based on diverse exemplars. The candidate responses are then aggregated using the same LLM to generate the final output. 

\subsection{Prompts with Diverse Exemplars}\label{subsec:prompt-ensembling}
 Traditional CoT based approaches rely on a single prompt comprised of a fixed set of exemplars. \cite{li-etal-2023-diverse-prompt} showed that constructing multiple prompts, by modifying the exemplars chosen for the purpose of In-Context-Learning (ICL), further enhances the reasoning capability of language models. On similar lines, we construct multiple LLM prompts by randomly sampling the exemplars for ICL multiple times using different seed settings. For each such LLM prompt, we generate a candidate response using Greedy Decoding.   
 
\subsection{LLM-based Aggregation}\label{subsec:aggregation}
 USC~\cite{Chen2023UniversalSF} that has been shown to accurately select the most consistent response among multiple SC responses using majority consensus. We follow USC and extract the final response from multiple candidate responses accordingly.

\section{Experiments}\label{sec:experiments}
\subsection{Dataset}
 We consider two publicly available datasets for the purpose of our experiments - 
\begin{itemize}
\item \textbf{SVAMP}~\cite{patel-etal-2021} Comprises of elementary-level Math Word Problems. Each problem consists of a short natural language narrative that describes a state of the world and poses a question about some unknown quantities.

\item \textbf{AI2 Reasoning Challenge (ARC)}~\cite{clark2018thinksolvedquestionanswering} is a multiple-choice question-answering dataset, containing questions from science exams from grade 3 to grade 9 and is further split in two partitions - `ARC-Easy' and `ARC-Challenge' where `ARC-Challenge' partition contains relatively more difficult questions that require reasoning
\end{itemize}

We report results on the validation split of each dataset. We restrict the ARC dataset to `ARC-Challenge' only and work with 30\% of the data sampled at random. Table~\ref{table:DatasetDetails} captures the corresponding details of the validation datasets considered for the experiments in the paper.

\begin{table}[h]
\centering
\begin{tabular}{|p{2.5cm}|p{4cm}|}
\hline
\textbf{Dataset Name} & \textbf{Number of Validation Samples}\\
\hline
SVAMP & 300 \\
\hline
ARC  & 345 \\\hline
\end{tabular}
\caption{Validation dataset size for SVAMP and ARC datasets}
\label{table:DatasetDetails}
\end{table}

\subsection{Baseline Strategies}
To benchmark our approach, PEDAL, we include the following baselines

\begin{itemize}
\item \textbf{Greedy Decoding} - We run the LLM to select the token with the highest probability at each step to generate the final output. 
\item \textbf{USC} - We run SC with CoT prompting and select the most consistent answer among all candidate responses using the same LLM.
\item \textbf{Unified Diverse Exemplars} - To understand the impact of multiple candidate responses generated in PEDAL using diverse prompts, we combine all such diverse exemplars directly into a single ICL prompt and run Greedy Decoding. We refer to this baseline as ``Unified Diverse Exemplars'' (UDE).

\end{itemize}

\subsection{Experiment Setting}
Each of the strategies were run using Qwen2-7B-Instruct~\cite{yang2024qwen2technicalreport} and Llama-3-8B-Instruct~\cite{Touvron2023LLaMAOA}. We measure the performance using accuracy and the number of output tokens. For purposes of reporting, we also share the number of input tokens consumed by the strategies. The LLMs were run using 4-bit quantization~\cite{dettmers2023qloraefficientfinetuningquantized}. Each experiment is run under three random seed settings for reproducibility. We pick three exemplars per experiment for the ICL prompt construction with each dataset. For each experiment, USC is run to generate three intermediate outputs and PEDAL is run with three diverse input prompts. 

\begin{table}[h]
\centering
\begin{tabular}{|p{1.2cm}|p{2.5cm}|p{2.5cm}|}
\hline
\textbf{Model} & \textbf{Approach} & \textbf{Accuracy}\\\hline
\multirow{4}{*}{Qwen2} & Greedy  & 76.0 $\pm$ 1.52 \\\cline{2-3}
 & \textbf{USC} & \textbf{80.33} $\pm$ \textbf{0.98} \\\cline{2-3}
 & UDE & 75.67 $\pm$ 0.0 \\\cline{2-3}
 & PEDAL  & 77.89 $\pm$  1.28 \\\hline
\multirow{4}{*}{Llama3} & Greedy & 70.22 $\pm$  1.03 \\\cline{2-3}
 & USC & 72.99 $\pm$ 0.47 \\\cline{2-3}
 & UDE & 70.67 $\pm$ 0.0 \\\cline{2-3}
 & \textbf{PEDAL} & \textbf{74.11} $\pm$  \textbf{0.57} \\\hline
\end{tabular}
\caption{Performance comparison of Greedy Decoding, USC, UDE and PEDAL for SVAMP dataset using Accuracy. Averaged scores across 3 seeds are reported along with the standard deviation. Best performing strategy per model has been highlighted in \textbf{bold}}
\label{tab:svamp-accuracy}
\end{table}

\begin{table}[t]
\centering
\begin{tabular}{|p{1.1cm}|p{2cm}|p{1.5cm}|p{1.5cm}|}
\hline
\textbf{Model} & \textbf{Approach} & \multicolumn{2}{|p{3cm}|}{\textbf{Token Count}}\\
\cline{3-4}
 & & \textbf{Input} & \textbf{Output} \\ \hline
\multirow{2}{*}{Qwen2} & USC  & 902.89 $\pm$ 2.16 & 502.75 $\pm$ 1.43 \\\cline{2-4}
 & \textbf{PEDAL}  & 1342.18 $\pm$ 86.87 & \textbf{191.99} $\pm$ \textbf{0.22} \\\hline
\multirow{2}{*}{Llama3} & USC  & 693.46 $\pm$ 8.79 & 923.56 $\pm$ 1.51 \\\cline{2-4}
 & \textbf{PEDAL}  & 1261.51 $\pm$ 64.95 & \textbf{197.72} $\pm$ \textbf{0.2} \\\hline
\end{tabular}
\caption{Performance comparison of USC and PEDAL for SVAMP dataset using the number of output tokens. Averaged counts across 3 seeds are reported along with the standard deviation. Best performing strategy per model has been highlighted in \textbf{bold}}
\label{tab:svamp-token_count}
\end{table}

\begin{table}[b]
\centering
\begin{tabular}{|p{1.2cm}|p{2.5cm}|p{2.5cm}|}
\hline
\textbf{Model} & \textbf{Approach} & \textbf{Accuracy} \\
\hline
\multirow{4}{*}{Qwen2} & Greedy & 83.38 $\pm$ 0.55 \\\cline{2-3}
 & \textbf{USC} & \textbf{84.35} $\pm$ \textbf{0.62} \\\cline{2-3}
 & UDE & 84.06 $\pm$ 0.0 \\\cline{2-3}
 & PEDAL  & 83.77 $\pm$ 0.47 \\\hline

\multirow{4}{*}{Llama3} & Greedy & 76.52 $\pm$ 1.44 \\\cline{2-3}
 & USC & 71.88 $\pm$ 0.71 \\\cline{2-3}
 & UDE & 76.52 $\pm$ 0.0 \\\cline{2-3}
 & \textbf{PEDAL}  & \textbf{78.55} $\pm$ \textbf{0.47} \\\hline
\end{tabular}
\caption{Performance comparison of greedy decoding, USC, UDE and PEDAL for ARC dataset using Accuracy. Averaged scores across 3 seeds are reported along with the standard deviation. Best performing strategy per model has been highlighted in \textbf{bold}}
\label{tab:arc-accuracy}
\end{table}

\begin{table}[b]
\centering
\begin{tabular}{|p{1.1cm}|p{2cm}|p{1.5cm}|p{1.5cm}|}
\hline
\textbf{Model} & \textbf{Approach} & \multicolumn{2}{|p{3cm}|}{\textbf{Token Count}}\\
\cline{3-4}
 & & \textbf{Input} & \textbf{Output} \\ \hline
\multirow{2}{*}{Qwen2} & USC  & 1153.04 $\pm$ 1.96 & 668.71 $\pm$ 7.19 \\\cline{2-4}
 & \textbf{PEDAL}  & 1179.76 $\pm$ 100.10 & \textbf{99.47} $\pm$ \textbf{10.05} \\\hline

\multirow{2}{*}{Llama3} & USC  & 1072.96 $\pm$ 5.67 & 928.1 $\pm$ 1.31 \\\cline{2-4}
 & \textbf{PEDAL}  & 1185.27 $\pm$ 115.08 & \textbf{196.83} $\pm$ \textbf{0.11} \\\hline
\end{tabular}
\caption{Performance comparison of USC and PEDAL for ARC dataset using the number of output tokens. Averaged counts across 3 seeds are reported along with the standard deviation. Best performing strategy per model has been highlighted in \textbf{bold}}
\label{tab:arc-token_count}
\end{table}
\section{Results and Analysis}\label{sec:analysis}

\begin{table}[h]
\centering
\begin{tabular}{|p{1.5cm}|p{2.5cm}|p{2.5cm}|}
\hline
\textbf{Number of Prompts} & \textbf{SVAMP} & \textbf{ARC}\\
\hline
2 & 77.0 $\pm$ 0.98 & 83.96 $\pm$ 0.36 \\\hline 
3 & 77.89 $\pm$  1.28 & 83.77 $\pm$ 0.47 \\\hline
4 & 78.22 $\pm$ 1.34  & 83.87 $\pm$ 0.49 \\\hline

\hline
\end{tabular}
\caption{Effect of number of prompts on performance using Qwen2 with SVAMP and ARC datasets. Averaged scores across 3 seeds are reported along with the standard deviation. }
\label{table:proposer-count}
\end{table}
Table~\ref{tab:svamp-accuracy} and Table~\ref{tab:svamp-token_count} show the performance metrics for different strategies using SVAMP dataset. Similarly, Table~\ref{tab:arc-accuracy} and Table~\ref{tab:arc-token_count} capture the performance metrics for the ARC dataset. We observe that our proposed approach consistently performs better than Greedy Decoding in terms of accuracy and outperforms USC in terms of the number of output tokens. 

\subsection{Arithmetic Reasoning}
As shown in Table~\ref{tab:svamp-accuracy}, PEDAL displays improvement over Greedy Decoding on the SVAMP dataset. With Qwen2, PEDAL achieves an average accuracy of 77.89\% while Greedy Decoding achieves an average accuracy of 76\% implying a 1.89\% improvement. PEDAL also outperforms UDE which achieves an accuracy of 75.67\%. USC achieves the accuracy of 80.33\%. Similarly, with Llama3, we observe that PEDAL achieves an average accuracy of 74.11\% while Greedy Decoding achieves a score of 70.22\% resulting in 3.89\% improvement. However, with Llama3, we observe that USC achieves an accuracy of 72.99\% which is lesser than PEDAL while UDE achieves an accuracy 70.67\% marginally outperforming Greedy Decoding.

As shown in Table~\ref{tab:svamp-token_count}, with Qwen2, USC processes approximately 903 input tokens and 503 output tokens while PEDAL processes 1,343 input tokens with 192 output tokens making our approach evidently more cost efficient. With Llama3, USC processes an average of 694 input tokens and 924 output tokens while PEDAL processes 1,262 input tokens and 198 output tokens. While USC relies on lesser input tokens than PEDAL, the cost of output tokens with USC is more than 4 times the output token cost with PEDAL making our approach more cost efficient. 

\subsection{Multiple-Choice Question Answering}
As shown in Table~\ref{tab:arc-accuracy}, the strategies show a similar relationship with experiments run on the ARC dataset. With Qwen2, PEDAL achieves a marginal improvement of 0.39\% over Greedy Decoding with an average accuracy of 83.77\% while Greedy Decoding has an average accuracy of 83.38\%. UDE outperforms PEDAL with an accuracy of 84.06\% while USC still achieves the best performance with an accuracy of 84.35\%. With Llama-3, PEDAL shows a 2.03\% improvement with a score of 78.55\% and greedy decoding achieves 76.52\%. UDE achieves an accuracy of 76.52\% matching the performance of Greedy Decoding. Surprisingly, USC achieves an accuracy of 71.88\% which is relatively the least among the strategies. With USC, the main goal of the paper is to benchmark the proposed approach in terms of token count. To prevent diverging from the primary focus area, we leave deeper analysis of this behaviour to future work.

As shown in Table~\ref{tab:arc-token_count}, with Qwen2, our approach outperforms USC where USC processes roughly 1,154 input tokens and 669 output tokens on an average while PEDAL processes 1,180 input tokens with 100 output tokens. With Llama3, USC processes 1,073 input tokens and 929 output tokens while PEDAL processes 1,186 input tokens and 197 output tokens. Our approach is the better choice in terms of the number of output tokens processed by the LLM.

\subsection{Comparison to CoT}
Similar to PEDAL, CoT has been shown to be more accurate than Greedy Decoding and less expensive in terms of inference compared to SC. Based on pre-liminary interpolation of the number of output tokens using Table~\ref{tab:svamp-token_count} and Table~\ref{tab:arc-token_count}, we compare the number of output tokens consumed in a single intermediate output in SC (equivalent to CoT) with the number of output tokens in PEDAL. With Llama3, we observe that PEDAL would be more cost efficient for both datasets. With Qwen2, we observe that PEDAL would be more cost efficient for the ARC dataset but may prove to be more expensive for the SVAMP dataset in comparison to CoT. While PEDAL seems to be more reliably consistent, it would be interesting to further investigate and arrive at definitive conclusions. We intend to evaluate the merits and drawbacks of both approaches in a practical setting in future work. 

\subsection{Impact of Number of Diverse Prompts}
We re-run the experiments for both datasets with our best performing model, Qwen2, by varying the number of prompts to study how it affects the performance. As shown in Table~\ref{table:proposer-count}, we additionally run the experiments for two and four diverse prompts under three seed settings. We observe slight improvements as we increase the number of prompts with the SVAMP dataset. However, we do not observe any such specific pattern with the ARC dataset.

\section{Conclusion}\label{sec:conclusion}
In this paper, we explored self-ensembling with LLMs using diverse exemplars with LLM based output aggregation. We observed that this combination can perform better than Greedy Decoding in terms of accuracy and achieve better cost efficiency than SC based methods. However, we restricted the experiments to small datasets that allowed benchmarking approaches using exact match without additional manual annotation efforts. In future work, we plan to explore possibilities on extending such ensembling strategies to a wider range of problem settings involving free-form text generation to further deep dive into strengths and weaknesses of our proposed system.

\bibliography{anthology,custom}

\begin{thebibliography}{45}
\expandafter\ifx\csname natexlab\endcsname\relax\def\natexlab#1{#1}\fi

\bibitem[{Arora et~al.(2022)Arora, Narayan, Chen, Orr, Guha, Bhatia, Chami, Sala, and Ré}]{arora2022askanythingsimplestrategy}
Simran Arora, Avanika Narayan, Mayee~F. Chen, Laurel Orr, Neel Guha, Kush Bhatia, Ines Chami, Frederic Sala, and Christopher Ré. 2022.
\newblock \href {http://arxiv.org/abs/2210.02441} {Ask me anything: A simple strategy for prompting language models}.

\bibitem[{Bach et~al.(2022)Bach, Sanh, Yong, Webson, Raffel, Nayak, Sharma, Kim, Bari, Fevry, Alyafeai, Dey, Santilli, Sun, Ben-david, Xu, Chhablani, Wang, Fries, Al-shaibani, Sharma, Thakker, Almubarak, Tang, Radev, Jiang, and Rush}]{bach-etal-2022}
Stephen Bach, Victor Sanh, Zheng~Xin Yong, Albert Webson, Colin Raffel, Nihal~V. Nayak, Abheesht Sharma, Taewoon Kim, M~Saiful Bari, Thibault Fevry, Zaid Alyafeai, Manan Dey, Andrea Santilli, Zhiqing Sun, Srulik Ben-david, Canwen Xu, Gunjan Chhablani, Han Wang, Jason Fries, Maged Al-shaibani, Shanya Sharma, Urmish Thakker, Khalid Almubarak, Xiangru Tang, Dragomir Radev, Mike Tian-jian Jiang, and Alexander Rush. 2022.
\newblock \href {https://doi.org/10.18653/v1/2022.acl-demo.9} {{P}rompt{S}ource: An integrated development environment and repository for natural language prompts}.
\newblock In \emph{Proceedings of the 60th Annual Meeting of the Association for Computational Linguistics: System Demonstrations}, pages 93--104, Dublin, Ireland. Association for Computational Linguistics.

\bibitem[{Brown et~al.(2020)Brown, Mann, Ryder, Subbiah, Kaplan, Dhariwal, Neelakantan, Shyam, Sastry, Askell, Agarwal, Herbert-Voss, Krueger, Henighan, Child, Ramesh, Ziegler, Wu, Winter, Hesse, Chen, Sigler, Litwin, Gray, Chess, Clark, Berner, McCandlish, Radford, Sutskever, and Amodei}]{brown2020GPT3}
Tom~B. Brown, Benjamin Mann, Nick Ryder, Melanie Subbiah, Jared Kaplan, Prafulla Dhariwal, Arvind Neelakantan, Pranav Shyam, Girish Sastry, Amanda Askell, Sandhini Agarwal, Ariel Herbert-Voss, Gretchen Krueger, Tom Henighan, Rewon Child, Aditya Ramesh, Daniel~M. Ziegler, Jeffrey Wu, Clemens Winter, Christopher Hesse, Mark Chen, Eric Sigler, Mateusz Litwin, Scott Gray, Benjamin Chess, Jack Clark, Christopher Berner, Sam McCandlish, Alec Radford, Ilya Sutskever, and Dario Amodei. 2020.
\newblock Language models are few-shot learners.
\newblock In \emph{Proceedings of the 34th International Conference on Neural Information Processing Systems}, NIPS'20, Red Hook, NY, USA. Curran Associates Inc.

\bibitem[{Cai et~al.(2024)Cai, Jiang, Wang, Tang, Kim, and Huang}]{cai2024surveymixtureexperts}
Weilin Cai, Juyong Jiang, Fan Wang, Jing Tang, Sunghun Kim, and Jiayi Huang. 2024.
\newblock \href {http://arxiv.org/abs/2407.06204} {A survey on mixture of experts}.

\bibitem[{Chase(2022)}]{Chase_LangChain_2022}
Harrison Chase. 2022.
\newblock \href {https://github.com/langchain-ai/langchain} {{LangChain}}.

\bibitem[{Chen et~al.(2024)Chen, Liu, Chen, Gu, Wu, Tao, Fu, and Ye}]{chen2024insidellmsinternalstates}
Chao Chen, Kai Liu, Ze~Chen, Yi~Gu, Yue Wu, Mingyuan Tao, Zhihang Fu, and Jieping Ye. 2024.
\newblock \href {http://arxiv.org/abs/2402.03744} {Inside: Llms' internal states retain the power of hallucination detection}.

\bibitem[{Chen et~al.(2023{\natexlab{a}})Chen, Zaharia, and Zou}]{chen2023frugalgptuselargelanguage}
Lingjiao Chen, Matei Zaharia, and James Zou. 2023{\natexlab{a}}.
\newblock \href {http://arxiv.org/abs/2305.05176} {Frugalgpt: How to use large language models while reducing cost and improving performance}.

\bibitem[{Chen et~al.(2023{\natexlab{b}})Chen, Aksitov, Alon, Ren, Xiao, Yin, Prakash, Sutton, Wang, and Zhou}]{Chen2023UniversalSF}
Xinyun Chen, Renat Aksitov, Uri Alon, Jie Ren, Kefan Xiao, Pengcheng Yin, Sushant Prakash, Charles Sutton, Xuezhi Wang, and Denny Zhou. 2023{\natexlab{b}}.
\newblock \href {https://api.semanticscholar.org/CorpusID:265498407} {Universal self-consistency for large language model generation}.
\newblock \emph{ArXiv}, abs/2311.17311.

\bibitem[{Cheng et~al.(2024)Cheng, Zhang, and Shi}]{cheng2024surveydeepneuralnetwork}
Hongrong Cheng, Miao Zhang, and Javen~Qinfeng Shi. 2024.
\newblock \href {http://arxiv.org/abs/2308.06767} {A survey on deep neural network pruning-taxonomy, comparison, analysis, and recommendations}.

\bibitem[{Chng(2024)}]{Chng2024OutputTokenGenCost}
Peter Chng. 2024.
\newblock \href {https://peterchng.com/blog/2024/05/01/why-do-llm-input-tokens-cost-less-than-output-tokens/} {Why do llm input tokens cost less than output tokens?}

\bibitem[{Chowdhery et~al.(2022)Chowdhery, Narang, Devlin, Bosma, Mishra, Roberts, Barham, Chung, Sutton, Gehrmann, Schuh, Shi, Tsvyashchenko, Maynez, Rao, Barnes, Tay, Shazeer, Prabhakaran, Reif, Du, Hutchinson, Pope, Bradbury, Austin, Isard, Gur-Ari, Yin, Duke, Levskaya, Ghemawat, Dev, Michalewski, Garc{\'i}a, Misra, Robinson, Fedus, Zhou, Ippolito, Luan, Lim, Zoph, Spiridonov, Sepassi, Dohan, Agrawal, Omernick, Dai, Pillai, Pellat, Lewkowycz, Moreira, Child, Polozov, Lee, Zhou, Wang, Saeta, D{\'i}az, Firat, Catasta, Wei, Meier-Hellstern, Eck, Dean, Petrov, and Fiedel}]{Chowdhery2022PaLMSL}
Aakanksha Chowdhery, Sharan Narang, Jacob Devlin, Maarten Bosma, Gaurav Mishra, Adam Roberts, Paul Barham, Hyung~Won Chung, Charles Sutton, Sebastian Gehrmann, Parker Schuh, Kensen Shi, Sasha Tsvyashchenko, Joshua Maynez, Abhishek Rao, Parker Barnes, Yi~Tay, Noam~M. Shazeer, Vinodkumar Prabhakaran, Emily Reif, Nan Du, Benton~C. Hutchinson, Reiner Pope, James Bradbury, Jacob Austin, Michael Isard, Guy Gur-Ari, Pengcheng Yin, Toju Duke, Anselm Levskaya, Sanjay Ghemawat, Sunipa Dev, Henryk Michalewski, Xavier Garc{\'i}a, Vedant Misra, Kevin Robinson, Liam Fedus, Denny Zhou, Daphne Ippolito, David Luan, Hyeontaek Lim, Barret Zoph, Alexander Spiridonov, Ryan Sepassi, David Dohan, Shivani Agrawal, Mark Omernick, Andrew~M. Dai, Thanumalayan~Sankaranarayana Pillai, Marie Pellat, Aitor Lewkowycz, Erica Moreira, Rewon Child, Oleksandr Polozov, Katherine Lee, Zongwei Zhou, Xuezhi Wang, Brennan Saeta, Mark D{\'i}az, Orhan Firat, Michele Catasta, Jason Wei, Kathleen~S. Meier-Hellstern, Douglas Eck, Jeff Dean, Slav Petrov,
  and Noah Fiedel. 2022.
\newblock \href {https://api.semanticscholar.org/CorpusID:247951931} {Palm: Scaling language modeling with pathways}.
\newblock \emph{J. Mach. Learn. Res.}, 24:240:1--240:113.

\bibitem[{Clark et~al.(2018)Clark, Cowhey, Etzioni, Khot, Sabharwal, Schoenick, and Tafjord}]{clark2018thinksolvedquestionanswering}
Peter Clark, Isaac Cowhey, Oren Etzioni, Tushar Khot, Ashish Sabharwal, Carissa Schoenick, and Oyvind Tafjord. 2018.
\newblock \href {http://arxiv.org/abs/1803.05457} {Think you have solved question answering? try arc, the ai2 reasoning challenge}.

\bibitem[{Dettmers et~al.(2023)Dettmers, Pagnoni, Holtzman, and Zettlemoyer}]{dettmers2023qloraefficientfinetuningquantized}
Tim Dettmers, Artidoro Pagnoni, Ari Holtzman, and Luke Zettlemoyer. 2023.
\newblock \href {http://arxiv.org/abs/2305.14314} {Qlora: Efficient finetuning of quantized llms}.

\bibitem[{Fernando et~al.(2023)Fernando, Banarse, Michalewski, Osindero, and Rocktäschel}]{fernando2023promptbreederselfreferentialselfimprovementprompt}
Chrisantha Fernando, Dylan Banarse, Henryk Michalewski, Simon Osindero, and Tim Rocktäschel. 2023.
\newblock \href {http://arxiv.org/abs/2309.16797} {Promptbreeder: Self-referential self-improvement via prompt evolution}.

\bibitem[{Gou et~al.(2021)Gou, Yu, Maybank, and Tao}]{Gou_2021}
Jianping Gou, Baosheng Yu, Stephen~J. Maybank, and Dacheng Tao. 2021.
\newblock \href {https://doi.org/10.1007/s11263-021-01453-z} {Knowledge distillation: A survey}.
\newblock \emph{International Journal of Computer Vision}, 129(6):1789–1819.

\bibitem[{Hou et~al.(2023)Hou, O'Connor, Andreas, Chang, and Zhang}]{Hou2023PromptBoosting}
Bairu Hou, Joe O'Connor, Jacob Andreas, Shiyu Chang, and Yang Zhang. 2023.
\newblock Promptboosting: black-box text classification with ten forward passes.
\newblock In \emph{Proceedings of the 40th International Conference on Machine Learning}, ICML'23. JMLR.org.

\bibitem[{Huang et~al.(2024)Huang, Lu, Chen, Wan, and Duan}]{huang2024enhancinglargelanguagemodels}
Baizhou Huang, Shuai Lu, Weizhu Chen, Xiaojun Wan, and Nan Duan. 2024.
\newblock \href {http://arxiv.org/abs/2309.17272} {Enhancing large language models in coding through multi-perspective self-consistency}.

\bibitem[{Jacob et~al.(2018)Jacob, Kligys, Chen, Zhu, Tang, Howard, Adam, and Kalenichenko}]{Jacob_2018_CVPR}
Benoit Jacob, Skirmantas Kligys, Bo~Chen, Menglong Zhu, Matthew Tang, Andrew Howard, Hartwig Adam, and Dmitry Kalenichenko. 2018.
\newblock Quantization and training of neural networks for efficient integer-arithmetic-only inference.
\newblock In \emph{Proceedings of the IEEE Conference on Computer Vision and Pattern Recognition (CVPR)}.

\bibitem[{Jiang et~al.(2023)Jiang, Ren, and Lin}]{jiang2023llmblenderensemblinglargelanguage}
Dongfu Jiang, Xiang Ren, and Bill~Yuchen Lin. 2023.
\newblock \href {http://arxiv.org/abs/2306.02561} {Llm-blender: Ensembling large language models with pairwise ranking and generative fusion}.

\bibitem[{Khattab et~al.(2023)Khattab, Singhvi, Maheshwari, Zhang, Santhanam, Vardhamanan, Haq, Sharma, Joshi, Moazam, Miller, Zaharia, and Potts}]{khattab2023dspycompilingdeclarativelanguage}
Omar Khattab, Arnav Singhvi, Paridhi Maheshwari, Zhiyuan Zhang, Keshav Santhanam, Sri Vardhamanan, Saiful Haq, Ashutosh Sharma, Thomas~T. Joshi, Hanna Moazam, Heather Miller, Matei Zaharia, and Christopher Potts. 2023.
\newblock \href {http://arxiv.org/abs/2310.03714} {Dspy: Compiling declarative language model calls into self-improving pipelines}.

\bibitem[{Li et~al.(2023{\natexlab{a}})Li, Peng, Zhang, Ding, Hu, and Shen}]{li2023deepmodelfusionsurvey}
Weishi Li, Yong Peng, Miao Zhang, Liang Ding, Han Hu, and Li~Shen. 2023{\natexlab{a}}.
\newblock \href {http://arxiv.org/abs/2309.15698} {Deep model fusion: A survey}.

\bibitem[{Li et~al.(2023{\natexlab{b}})Li, Lin, Zhang, Fu, Chen, Lou, and Chen}]{li-etal-2023-diverse-prompt}
Yifei Li, Zeqi Lin, Shizhuo Zhang, Qiang Fu, Bei Chen, Jian-Guang Lou, and Weizhu Chen. 2023{\natexlab{b}}.
\newblock \href {https://doi.org/10.18653/v1/2023.acl-long.291} {Making language models better reasoners with step-aware verifier}.
\newblock In \emph{Proceedings of the 61st Annual Meeting of the Association for Computational Linguistics (Volume 1: Long Papers)}, pages 5315--5333, Toronto, Canada. Association for Computational Linguistics.

\bibitem[{Li et~al.(2024)Li, Wang, Ding, and Chen}]{li2024largelanguagemodelsfinance}
Yinheng Li, Shaofei Wang, Han Ding, and Hang Chen. 2024.
\newblock \href {http://arxiv.org/abs/2311.10723} {Large language models in finance: A survey}.

\bibitem[{Long(2023)}]{long2023largelanguagemodelguided}
Jieyi Long. 2023.
\newblock \href {http://arxiv.org/abs/2305.08291} {Large language model guided tree-of-thought}.

\bibitem[{Lu et~al.(2022)Lu, Bartolo, Moore, Riedel, and Stenetorp}]{lu-etal-2022-fantastically-pe}
Yao Lu, Max Bartolo, Alastair Moore, Sebastian Riedel, and Pontus Stenetorp. 2022.
\newblock \href {https://doi.org/10.18653/v1/2022.acl-long.556} {Fantastically ordered prompts and where to find them: Overcoming few-shot prompt order sensitivity}.
\newblock In \emph{Proceedings of the 60th Annual Meeting of the Association for Computational Linguistics (Volume 1: Long Papers)}, pages 8086--8098, Dublin, Ireland. Association for Computational Linguistics.

\bibitem[{Patel et~al.(2021)Patel, Bhattamishra, and Goyal}]{patel-etal-2021}
Arkil Patel, Satwik Bhattamishra, and Navin Goyal. 2021.
\newblock \href {https://doi.org/10.18653/v1/2021.naacl-main.168} {Are {NLP} models really able to solve simple math word problems?}
\newblock In \emph{Proceedings of the 2021 Conference of the North American Chapter of the Association for Computational Linguistics: Human Language Technologies}, pages 2080--2094, Online. Association for Computational Linguistics.

\bibitem[{Pitis et~al.(2023)Pitis, Zhang, Wang, and Ba}]{pitis2023boostedpromptensembleslarge}
Silviu Pitis, Michael~R. Zhang, Andrew Wang, and Jimmy Ba. 2023.
\newblock \href {http://arxiv.org/abs/2304.05970} {Boosted prompt ensembles for large language models}.

\bibitem[{Raffel et~al.(2020)Raffel, Shazeer, Roberts, Lee, Narang, Matena, Zhou, Li, and Liu}]{Raffel2020T5}
Colin Raffel, Noam Shazeer, Adam Roberts, Katherine Lee, Sharan Narang, Michael Matena, Yanqi Zhou, Wei Li, and Peter~J. Liu. 2020.
\newblock Exploring the limits of transfer learning with a unified text-to-text transformer.
\newblock \emph{J. Mach. Learn. Res.}, 21(1).

\bibitem[{Ratner et~al.(2017)Ratner, Bach, Ehrenberg, Fries, Wu, and R\'{e}}]{WeakSupervisionSnorkel2017}
Alexander Ratner, Stephen~H. Bach, Henry Ehrenberg, Jason Fries, Sen Wu, and Christopher R\'{e}. 2017.
\newblock \href {https://doi.org/10.14778/3157794.3157797} {Snorkel: rapid training data creation with weak supervision}.
\newblock \emph{Proc. VLDB Endow.}, 11(3):269–282.

\bibitem[{Schapire(2013)}]{schapire2013explaining}
Robert~E Schapire. 2013.
\newblock Explaining adaboost.
\newblock In \emph{Empirical inference}, pages 37--52. Springer.

\bibitem[{Shazeer(2019)}]{shazeer2019fasttransformerdecodingwritehead}
Noam Shazeer. 2019.
\newblock \href {http://arxiv.org/abs/1911.02150} {Fast transformer decoding: One write-head is all you need}.

\bibitem[{Singh et~al.(2023)Singh, Morris, Rush, Gao, and Deng}]{singh-etal-2023-tree}
Chandan Singh, John Morris, Alexander Rush, Jianfeng Gao, and Yuntian Deng. 2023.
\newblock \href {https://doi.org/10.18653/v1/2023.emnlp-main.384} {Tree prompting: Efficient task adaptation without fine-tuning}.
\newblock In \emph{Proceedings of the 2023 Conference on Empirical Methods in Natural Language Processing}, pages 6253--6267, Singapore. Association for Computational Linguistics.

\bibitem[{Touvron et~al.(2023)Touvron, Lavril, Izacard, Martinet, Lachaux, Lacroix, Rozi{\`e}re, Goyal, Hambro, Azhar, Rodriguez, Joulin, Grave, and Lample}]{Touvron2023LLaMAOA}
Hugo Touvron, Thibaut Lavril, Gautier Izacard, Xavier Martinet, Marie-Anne Lachaux, Timoth{\'e}e Lacroix, Baptiste Rozi{\`e}re, Naman Goyal, Eric Hambro, Faisal Azhar, Aurelien Rodriguez, Armand Joulin, Edouard Grave, and Guillaume Lample. 2023.
\newblock \href {https://api.semanticscholar.org/CorpusID:257219404} {Llama: Open and efficient foundation language models}.
\newblock \emph{ArXiv}, abs/2302.13971.

\bibitem[{Wang et~al.(2022)Wang, Wei, Schuurmans, Le, hsin Chi, and Zhou}]{Wang2022SelfConsistencyIC}
Xuezhi Wang, Jason Wei, Dale Schuurmans, Quoc Le, Ed~Huai hsin Chi, and Denny Zhou. 2022.
\newblock \href {https://api.semanticscholar.org/CorpusID:247595263} {Self-consistency improves chain of thought reasoning in language models}.
\newblock \emph{ArXiv}, abs/2203.11171.

\bibitem[{Wei et~al.(2022)Wei, Wang, Schuurmans, Bosma, hsin Chi, Xia, Le, and Zhou}]{Wei2022ChainOT}
Jason Wei, Xuezhi Wang, Dale Schuurmans, Maarten Bosma, Ed~Huai hsin Chi, F.~Xia, Quoc Le, and Denny Zhou. 2022.
\newblock \href {https://api.semanticscholar.org/CorpusID:246411621} {Chain of thought prompting elicits reasoning in large language models}.
\newblock \emph{ArXiv}, abs/2201.11903.

\bibitem[{Wu et~al.(2024)Wu, Liu, Gong, Wang, Li, Wang, Cai, and Zhao}]{wu2024paralleldecodinghiddentransfer}
Pengfei Wu, Jiahao Liu, Zhuocheng Gong, Qifan Wang, Jinpeng Li, Jingang Wang, Xunliang Cai, and Dongyan Zhao. 2024.
\newblock \href {http://arxiv.org/abs/2404.12022} {Parallel decoding via hidden transfer for lossless large language model acceleration}.

\bibitem[{Yang et~al.(2024)Yang, Yang, Hui, Zheng, Yu, Zhou, Li, Li, Liu, Huang, Dong, Wei, Lin, Tang, Wang, Yang, Tu, Zhang, Ma, Yang, Xu, Zhou, Bai, He, Lin, Dang, Lu, Chen, Yang, Li, Xue, Ni, Zhang, Wang, Peng, Men, Gao, Lin, Wang, Bai, Tan, Zhu, Li, Liu, Ge, Deng, Zhou, Ren, Zhang, Wei, Ren, Liu, Fan, Yao, Zhang, Wan, Chu, Liu, Cui, Zhang, Guo, and Fan}]{yang2024qwen2technicalreport}
An~Yang, Baosong Yang, Binyuan Hui, Bo~Zheng, Bowen Yu, Chang Zhou, Chengpeng Li, Chengyuan Li, Dayiheng Liu, Fei Huang, Guanting Dong, Haoran Wei, Huan Lin, Jialong Tang, Jialin Wang, Jian Yang, Jianhong Tu, Jianwei Zhang, Jianxin Ma, Jianxin Yang, Jin Xu, Jingren Zhou, Jinze Bai, Jinzheng He, Junyang Lin, Kai Dang, Keming Lu, Keqin Chen, Kexin Yang, Mei Li, Mingfeng Xue, Na~Ni, Pei Zhang, Peng Wang, Ru~Peng, Rui Men, Ruize Gao, Runji Lin, Shijie Wang, Shuai Bai, Sinan Tan, Tianhang Zhu, Tianhao Li, Tianyu Liu, Wenbin Ge, Xiaodong Deng, Xiaohuan Zhou, Xingzhang Ren, Xinyu Zhang, Xipin Wei, Xuancheng Ren, Xuejing Liu, Yang Fan, Yang Yao, Yichang Zhang, Yu~Wan, Yunfei Chu, Yuqiong Liu, Zeyu Cui, Zhenru Zhang, Zhifang Guo, and Zhihao Fan. 2024.
\newblock \href {http://arxiv.org/abs/2407.10671} {Qwen2 technical report}.

\bibitem[{Yao et~al.(2023)Yao, Yu, Zhao, Shafran, Griffiths, Cao, and Narasimhan}]{yao2023treethoughtsdeliberateproblem}
Shunyu Yao, Dian Yu, Jeffrey Zhao, Izhak Shafran, Thomas~L. Griffiths, Yuan Cao, and Karthik Narasimhan. 2023.
\newblock \href {http://arxiv.org/abs/2305.10601} {Tree of thoughts: Deliberate problem solving with large language models}.

\bibitem[{Yu et~al.(2022)Yu, Quartey, and Schilder}]{yu2022legalpromptingteachinglanguage}
Fangyi Yu, Lee Quartey, and Frank Schilder. 2022.
\newblock \href {http://arxiv.org/abs/2212.01326} {Legal prompting: Teaching a language model to think like a lawyer}.

\bibitem[{Zhang et~al.(2023)Zhang, Liu, Wang, Wang, Sun, Wang, and Cai}]{zhang2023preferpromptensemblelearning}
Chenrui Zhang, Lin Liu, Jinpeng Wang, Chuyuan Wang, Xiao Sun, Hongyu Wang, and Mingchen Cai. 2023.
\newblock \href {http://arxiv.org/abs/2308.12033} {Prefer: Prompt ensemble learning via feedback-reflect-refine}.

\bibitem[{Zhao et~al.(2023)Zhao, Zhou, Li, Tang, Wang, Hou, Min, Zhang, Zhang, Dong, Du, Yang, Chen, Chen, Jiang, Ren, Li, Tang, Liu, Liu, Nie, and Wen}]{zhao2023surveylargelanguagemodels}
Wayne~Xin Zhao, Kun Zhou, Junyi Li, Tianyi Tang, Xiaolei Wang, Yupeng Hou, Yingqian Min, Beichen Zhang, Junjie Zhang, Zican Dong, Yifan Du, Chen Yang, Yushuo Chen, Zhipeng Chen, Jinhao Jiang, Ruiyang Ren, Yifan Li, Xinyu Tang, Zikang Liu, Peiyu Liu, Jian-Yun Nie, and Ji-Rong Wen. 2023.
\newblock \href {http://arxiv.org/abs/2303.18223} {A survey of large language models}.

\bibitem[{Zheng et~al.(2024)Zheng, Ning, Wang, Zhang, Zheng, Ye, and Chen}]{zheng2024surveylargelanguagemodelsCode}
Zibin Zheng, Kaiwen Ning, Yanlin Wang, Jingwen Zhang, Dewu Zheng, Mingxi Ye, and Jiachi Chen. 2024.
\newblock \href {http://arxiv.org/abs/2311.10372} {A survey of large language models for code: Evolution, benchmarking, and future trends}.

\bibitem[{Zhou et~al.(2022)Zhou, Scharli, Hou, Wei, Scales, Wang, Schuurmans, Bousquet, Le, and hsin Chi}]{Zhou2022LeasttoMostPE}
Denny Zhou, Nathanael Scharli, Le~Hou, Jason Wei, Nathan Scales, Xuezhi Wang, Dale Schuurmans, Olivier Bousquet, Quoc Le, and Ed~Huai hsin Chi. 2022.
\newblock \href {https://api.semanticscholar.org/CorpusID:248986239} {Least-to-most prompting enables complex reasoning in large language models}.
\newblock \emph{ArXiv}, abs/2205.10625.

\bibitem[{Zhou et~al.(2024)Zhou, Liu, Gu, Zou, Huang, Wu, Li, Chen, Zhou, Liu, Hua, Mao, You, Wu, Zheng, Clifton, Li, Luo, and Clifton}]{zhou2024LLMSurveyMedocine}
Hongjian Zhou, Fenglin Liu, Boyang Gu, Xinyu Zou, Jinfa Huang, Jinge Wu, Yiru Li, Sam~S. Chen, Peilin Zhou, Junling Liu, Yining Hua, Chengfeng Mao, Chenyu You, Xian Wu, Yefeng Zheng, Lei Clifton, Zheng Li, Jiebo Luo, and David~A. Clifton. 2024.
\newblock \href {http://arxiv.org/abs/2311.05112} {A survey of large language models in medicine: Progress, application, and challenge}.

\bibitem[{Zhu et~al.(2024)Zhu, Li, Liu, Ma, and Wang}]{zhu2024surveymodelcompressionlarge}
Xunyu Zhu, Jian Li, Yong Liu, Can Ma, and Weiping Wang. 2024.
\newblock \href {http://arxiv.org/abs/2308.07633} {A survey on model compression for large language models}.

\end{thebibliography}
\appendix

\end{document}